\def\year{2022}\relax
\documentclass[letterpaper]{article} % DO NOT CHANGE THIS
\usepackage{aaai22}  % DO NOT CHANGE THIS
\usepackage{times}  % DO NOT CHANGE THIS
\usepackage{helvet}  % DO NOT CHANGE THIS
\usepackage{courier}  % DO NOT CHANGE THIS
\usepackage[hyphens]{url}  % DO NOT CHANGE THIS
\usepackage{graphicx} % DO NOT CHANGE THIS
\usepackage{natbib}  % DO NOT CHANGE THIS AND DO NOT ADD ANY OPTIONS TO IT
\usepackage{caption} % DO NOT CHANGE THIS AND DO NOT ADD ANY OPTIONS TO IT
\DeclareCaptionStyle{ruled}{labelfont=normalfont,labelsep=colon,strut=off} % DO NOT CHANGE THIS
\usepackage{algorithm}
\usepackage{algorithmic}

\usepackage{newfloat}
\usepackage{listings}
\floatstyle{ruled}
\newfloat{listing}{tb}{lst}{}
\floatname{listing}{Listing}

\usepackage{amssymb, amsmath}
\usepackage{graphicx}
\usepackage{hyperref}
\usepackage{cleveref}
\usepackage{amsthm}

\newcommand{\doo}{\textrm{do}}

\newcommand{\prob}{ P }

\title{Probability trees and the value of a single intervention}
\author {
	Tue Herlau%\textsuperscript{\rm 1}
}
\affiliations {
	Technical University of Denmark, 2800 Lyngby, Denmark\\
	tuhe@dtu.dk%, ralars@dtu.dk
}

\begin{document}

\maketitle

\begin{abstract}
The most fundamental problem in statistical causality is determining causal relationships from limited data. Probability trees, which combine prior causal structures with Bayesian updates, 
have been suggested as a possible solution.	
In this work, we quantify the information gain from a single intervention and show that both the anticipated information gain, prior to making an intervention, and the expected gain from an intervention have simple expressions. This results in an active-learning method that simply selects the intervention with the highest anticipated gain, which we illustrate through several examples. Our work demonstrates how probability trees, and Bayesian estimation of their parameters, offer a simple yet viable approach to fast causal induction.
\end{abstract}

\noindent 
\section{Introduction}

Perhaps the most fundamental problem in statistical causality is the problem of \emph{causal induction} itself, namely how to obtain the general causal relationships from limited observations~\citep{griffiths2007mere}.

Humans are able to infer causal relationships from samples that are too small for statistical tests to produce significant results~\cite{gopnik2001causal}. This ability is seemingly at odds with assumptions that are at the center of many popular algorithms for learning causal graphical models, which often require statistical testing of independence claims using large samples of observations~\cite{spirtes2000causation,pearl2000models,janzing2012information}. % a notoriously data-hungry task. %which requires large samples of observations.

Bayesian causal induction and probability trees offer an intuitively appealing possible answer to the question of how humans can learn rapidly from such small samples, namely as a combination of prior knowledge in the form of observations and Bayesian updating~\cite{griffiths2009theory,ortega2015subjectivity}. This is accomplished by encoding alternative causal hypothesis directly in the probability tree, thereby making no distinction between random variables representing data and the causal hypothesis itself~\cite{genewein2020algorithms,shafer1996art}. %, thereby allowing us to predict the effect of interventions and distinguish between 

In this work, we show how probability trees and information theory can predict circumstances under which causal induction will occur quickly. We do this by defining the expected information gain of an intervention, and demonstrating how it can be estimated prior to performing the intervention, thereby leading to a natural active-learning method for causal induction. 

We will focus on arguably the most basic form of causal learning (and certainly the problem which has garnered the most attention in psychology~\cite{gopnik2001causal}), namely learning a single causal relationship $X \rightarrow Y$ or $Y \rightarrow X$ between two variables $X \in \{1,\dots,K_X\}$ and $Y \in \{1,\dots,K_Y\}$ in the very small sample limit.

\begin{figure}[t!]
	\centering
	\includegraphics[width=.75\linewidth]{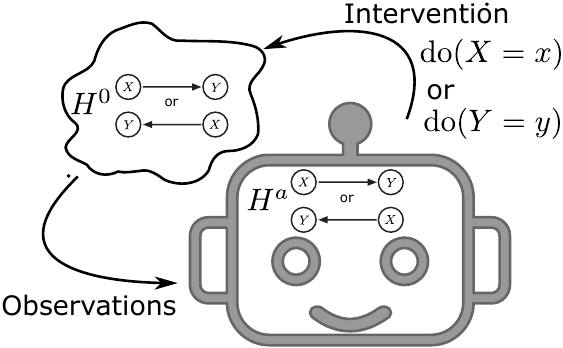}
	\caption{Either $X$ is the cause of $Y$ or $Y$ is the cause of $X$, and we distinguish between the Agents hypothesis $H^a$ about the causal orientation and the actual orientation $H^0$. Our overall goal is to quantify how much information can be gained about $H^0$ on average from a single intervention.}\label{fig1svg}
\end{figure}

We focus on two questions: (i) how much information will the agent \emph{anticipate} from an intervention? and (ii) how much information will an agent actually \emph{gain} from an intervention?. To quantify these expressions we distinguish between the agent's belief about the causal orientation $H^a \in \{h^a_{X \rightarrow Y}, h^a_{Y \rightarrow X}\}$ and the true causal orientation governing the world $H^0 \in \{h^0_{X \rightarrow Y}, h^0_{Y \rightarrow X} \}$ (see \cref{fig1svg}). This is necessary, since a reduction in uncertainty about $H^a$ is by itself meaningless, and we want to quantify the information gain about the true orientation $H^0$ \emph{regardless} of what it is. 

While both $H^0$ and $H^a$ have simple expressions, the anticipated information gain is a Jeffrey divergence.
  Intuitively, what makes an intervention $\doo(X=x)$ appealing is that there are $Y=y$-values such that $P(y|x) \neq P(y)$, however with \emph{rare} events being more informative than common events. 
 Priors will therefore be expected to play a more significant role in causal induction, a point we return to in Example 3. 
  Simulated examples\footnote{Code to reproduce all plots in this paper can be found at \url{https://gitlab.compute.dtu.dk/tuhe/probability_tree_learning}} show that  the gains can be estimated from small samples.

\paragraph{Related work}
The effect of a cause has been quantified using information theory~\cite{wieczorek2019information}, however, without considering learning in a Bayesian setting. Entropic causal inference (see \cite{compton2021entropic}) specifies circumstances where the causal direction between categorical variables can be determined from observational data under assumptions of limited entropy. Information geometry has been proposed as a means to infer causal orientation by relying on distributional assumptions~\cite{janzing2012information}, however both settings are different from the Bayesian learning setting considered here. Targeted intervention selection has also been considered by~\cite{agrawal2019abcd,tong2001active} but using a different measure to quantify the goodness of an intervention.

\section{Methods} 

\begin{figure}[t!]
	\centering
	\includegraphics[width=.85\linewidth]{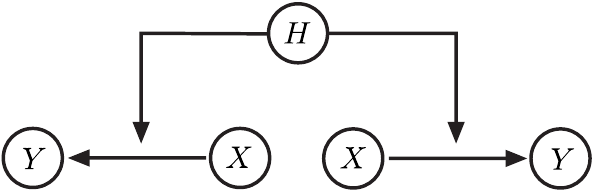}
	\caption{A probability tree for the problem involving $X$ and $Y$. The variable $H = h$ or $H = \neg h$ determines the causal orientation of $X$ and $Y$. Thus, given $H$, the effect of interventions is computed from the correct branch. }\label{fig2svg}
\end{figure}

A probability tree describes the relationships between the causal hypothesis $H$ and all variables (in this case $X$ and $Y$). In the tree, each internal node represents a causal mechanism, such that a path from the root node to a leaf corresponds to a realization of a causal mechanism~\cite{ortega2015subjectivity}.

The resolution order of the variables will typically depend on the path taken in the tree. In the left-most branch, $X$ precedes $Y$ under the hypothesis $h_{X\rightarrow Y}$, whereas in the right-most branch $Y$ precedes $X$. Thus, the probability tree allows the agent to model different potential mechanisms for the data-generating procedure. 

Using Pearls shorthand $\hat x = \doo(X=x)$ for an intervention, we can easily predict the effect of an intervention on $y$ from their joint distribution $P(X,Y)$. For instance, $P(y | \hat x, h_{X \rightarrow Y}) = P(p |x)$~\cite{genewein2020algorithms}. Therefore, we can compute the posterior probability of $h$ given an intervention $\hat x$ resulted in observing $Y=y$: % this mechanism: % we can easily understand this mechanism:

\begin{align}
	\prob(h|\hat{x},y)
	&= \frac{\prob(y|h,\hat{x})\prob(\hat{x}|h)\prob(h)}
	{ \sum_{h \in \{ h_{X \rightarrow Y}, h_{Y\rightarrow X}\} } \prob(y|h,\hat{x})\prob(\hat{x}|h)\prob(h) }\nonumber  \\  
	& = \frac{P(y | x)P(h) }{ P(y|x)P(h) + P(y  ) P(\neg h) } \label{eq:1}
\end{align}
It is thus clearly evident that interventions change the posterior belief about $H$ as long as $P(y | x) \neq p(y)$, but that purely observational data does not~\cite{ortega2015subjectivity}.

\subsection{Bayesian learning}
Let's now consider the case in which the probabilities such as $P(y | \hat x, h)$ are not known, but must be inferred from data. 

The data will be given as a $K_X \times K_Y$ matrix $n$ such that $n_{ij}$ is the number of times $(X = i, Y=j)$ has been jointly observed. For clarity, we will consistently use $Q$ to denote probabilities that are learned (from $n$) and $P$ for their actual value, as dictated by whatever mechanism governs the world. In that case, the Agent's belief about the causal orientation $H^a$, conditional on $n$, is given in \cref{eq:1} as long as we condition all probabilities on $n$. In this case, the left-hand side is $Q(h^a | \hat x, y, n)$ and we therefore need to compute expressions such as:
\begin{align}
Q(h  | n), \quad Q(y  | h, \hat x, n), \quad Q(y  | n).
\end{align}

To specify a suitable prior, we assume that observational data alone should remain uninformative about $h$, i.e. that $Q(h | x, y, n) = Q(h)$\footnote{This assumption is not innocent, as it is violated in work which assumes the causal orientation can be identified from observational data~ \cite{compton2021entropic}. This could be implemented by specifying ($P(x|y, h)$, $P(y|h)$ $P(x|y, h)$) and ($P(x|y, \neg h)$, $P(x|\neg h)$) separately.}. To ensure this, we must have
\begin{align}
Q(y | h, x, n)Q(x|n, h) = Q(y | \neg h, x, n)Q(x|n, \neg h) 
\end{align}
The natural way to ensure this is to specify the prior $Q(x,y|h)$ without reference to $h$, and then define the above expressions as the marginals and conditionals of the posterior. Specifically, we define

\begin{align*}
(\theta_{x,y} )_{x,y=1}^{K_X,K_Y} & \sim \textrm{Dir}( (\theta)_{xy} | \alpha 1_{K_X K_Y} ), \quad
Q(x,y | h)  = \theta_{x,y}.
\end{align*}
A standard application of Bayes' theorem gives us the standard posterior as $Q_n(x,y) = Q(x,y|n) = \frac{n_{xy}+\alpha}{ N + K_X K_Y\alpha  }$ from which all relevant quantities follow. For instance:
\begin{subequations}\label{eq:6}
\begin{align}
Q(y | h_{X \rightarrow y}, n, \hat x) & = Q_n(y|x) = \frac{ n_{yx} +\alpha }{ \sum_{y} n_{xy} + K_Y \alpha } \\ 
Q(y | h_{Y \rightarrow x}, n, \hat x) & = Q_n(y) = \frac{ \sum_{x} n_{xy} + K_X \alpha }{ N + K_X K_Y \alpha }.
\end{align}
\end{subequations}
	
	\begin{figure*}[t!]
		\centering
		\includegraphics[width=.33\linewidth]{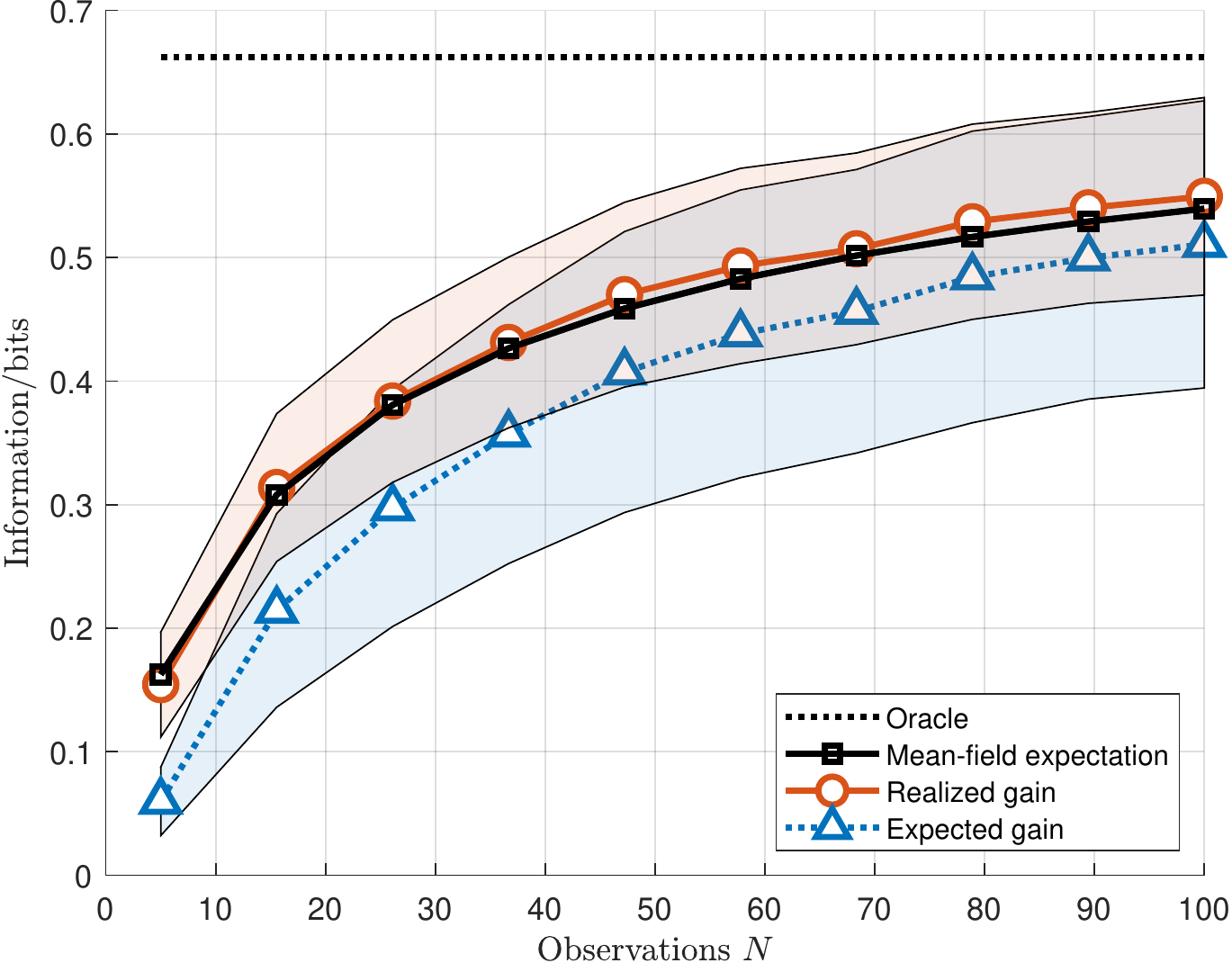}~
		\includegraphics[width=.33\linewidth]{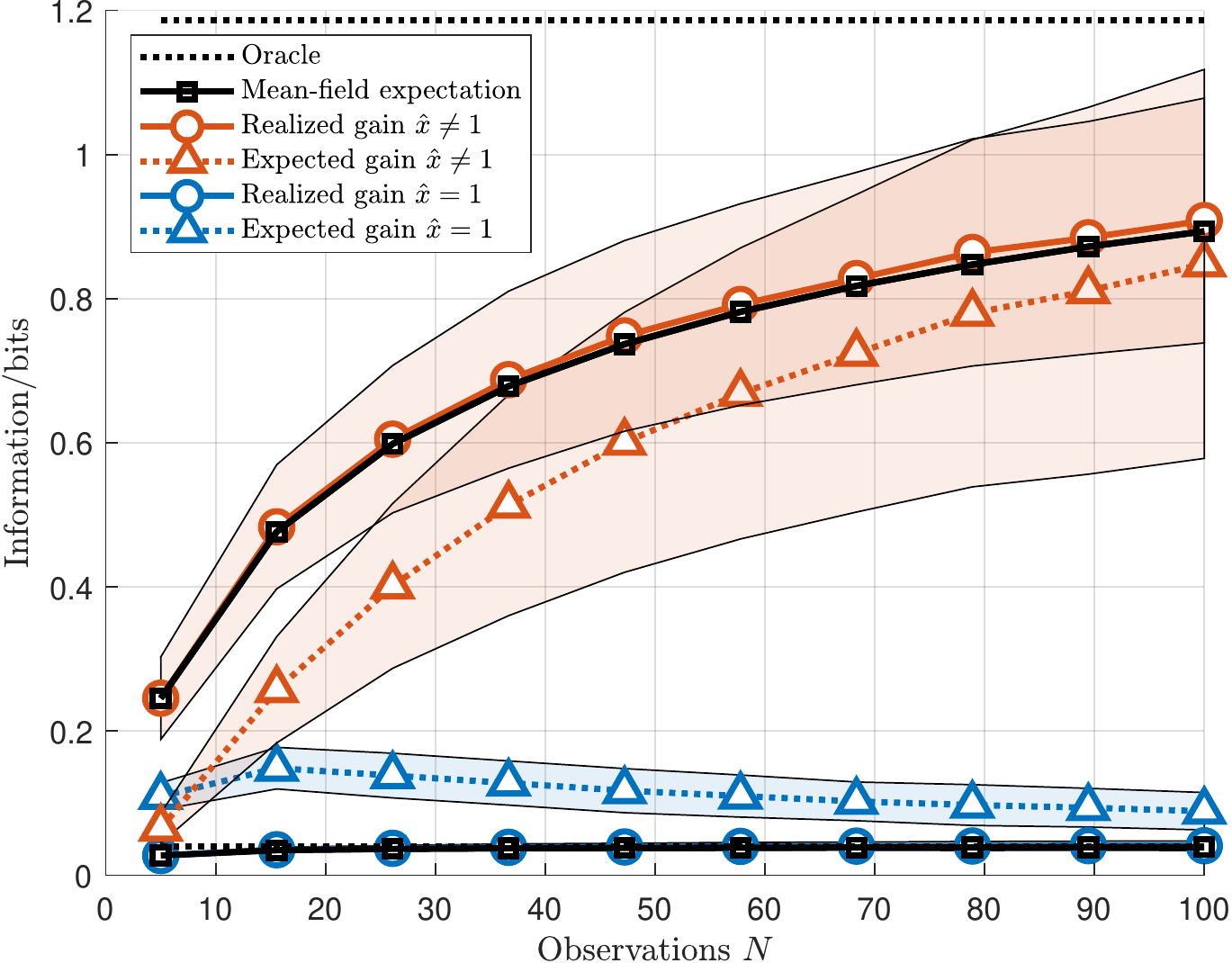}~
		\includegraphics[width=.33\linewidth]{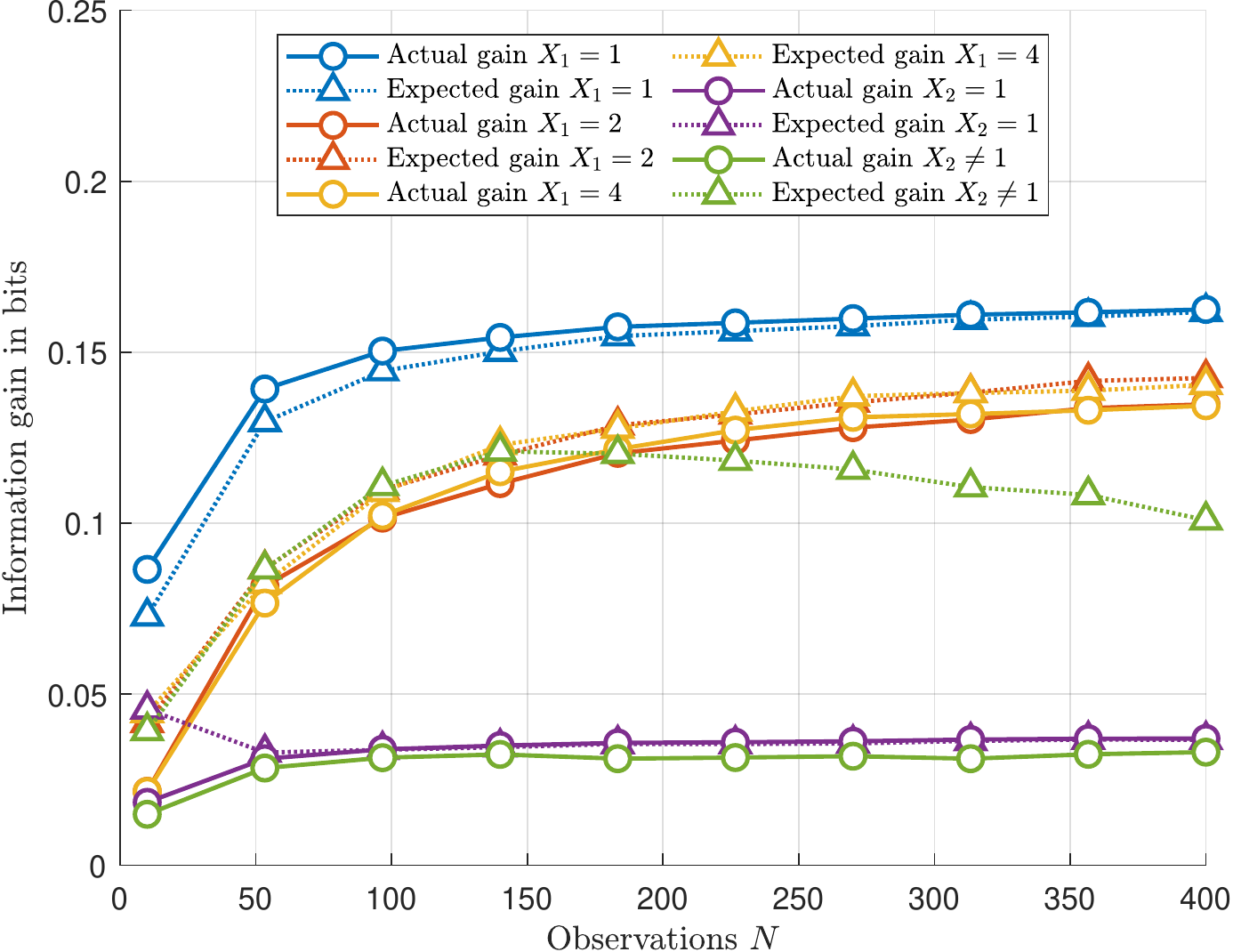}		
		\caption{The plots show both the expected information gain (triangles) and the realized gain (circles) for Example 1 (left), Example 2 (middle) and Example 3 (right). The expected gain, \cref{eq:9}, signifies the agent's estimate of the information gain from a given intervention $\hat x$ or $\hat y$, i.e. the agent will decide based on this value (by symmetry, all possible interventions are represented in the figures). The realized gain (computed using  \cref{eq:12}), is the gain the agent will actually obtain. Each simulation is run 1000 times, using a varying number of observations $N$, and the shaded region indicates one standard deviation. }\label{fig3}
	\end{figure*}

\subsection{Information gain}
If we suppose the agent performs an intervention $\doo(X=x)$ and observes $y$, and assume the agent is a-priori uninformed $P(H^a=h) = \frac{1}{2}$, then the change in the agent's belief about the causal orientation $h^a_{X\rightarrow Y}$, measured in bits, is given by \cref{eq:1} and \cref{eq:6}: %amount of evidence gained about either of the two causal hypothesis is: 
\begin{align}
I(\hat x | y) & = 
\log \frac{Q( h^a_{X \rightarrow Y} | n, \hat x, y ) }{ Q( h^a_{Y \rightarrow X} | n, \hat x, y ) } 
 = \log \frac{ Q(y | \hat x, n, h^a_{X\rightarrow Y} ) }
{ Q(y | \hat x, n, h^a_{ Y \rightarrow X} ) } \nonumber \\
& = \log \frac{ Q_n(y | x)  }
{ Q_n(y ) }.  \label{eq:6}
\end{align}

Recall that the true causal orientation is $H^0$. When $H^0 = h^0_{X \rightarrow Y}$, the above expression captures the gain in evidence for $H^0$, otherwise if $H^0 = h^0_{Y \rightarrow X}$ the gain in evidence for the \emph{true} direction is $-I(\hat x | y)$. Thus, the information gain in the \emph{true} direction is
\begin{align}
\Delta_{ \doo(X=x) } & = \sum_{h_0} P(h^0) \sum_{y | h^0} P(y | h^0, \hat x ) (-1)^{ \delta_{h^0, h^0_{X \rightarrow Y} } - 1 } I(\hat x | y) \nonumber \\
& = \frac{1}{2} \sum_{y} \left[P(y | x) - P(y)   \right]  I(\hat x | y).   \label{eq:12}
\end{align}
We refer to this as the \emph{realized information gain}, as it represents the change in evidence for the true orientation the agent would \emph{in fact} experience after performing $\hat x$. 

Prior to the intervention $\hat x$, the agent can reason about the expected information gain. This can be computed by simply inserting $Q$ instead of $P$ in \cref{eq:12}. Upon simplification, we see that the information gain is the Jeffrey divergence $D_J(p,q) = \sum_{x} (p(x)-q(x))\log \frac{p(x)}{q(x)}$
\begin{align}
\Delta^\text{Expected}_{\doo(X=x)} = D_J( Q_n(Y | x), Q_n(Y) ).
\label{eq:9}
\end{align}
We refer to this as the \emph{expected} information gain, as it is the gained information about the true causal orientation the agent expects prior to performing $\hat x$.

\subsubsection{Example 1: Two correlated variables}
We first consider the simplest possible case where $K_X = K_Y = 2$, the true causal orientation is $H^0 = h^0_{X\rightarrow Y}$, and $P(x,y)=P_{xy}$, considered as a $2 \times 2$ matrix, is
$ P = \begin{bmatrix}
	\rho  & (1-\rho) \\
	(1-\rho) & \rho 
\end{bmatrix} $
 where $\rho$ controls the correlation of $X$ and $Y$. We first consider a \emph{mean-field case} where $n = \frac{N}{2}P$. The realized information gain will be given by \cref{eq:12}

\begin{align}
\Delta_{\doo(X=x) } & = \frac{1}{2}
(\rho-\frac{1}{2} ) \log \frac{ N \rho + 2\alpha }{ N(1-\rho) + 2\alpha }. %+  \right. \\
\end{align}
As expected, in the uncorrelated case $\rho= \frac{1}{2}$ the information gain is 0. On the other hand, if the problem is deterministic $\rho = 1$ the information gain is $\Delta_{\doo(X=x) } = \frac{1}{4}\log \frac{ N + 2\alpha }{ 2\alpha}$. In the case of two variables, the expected amount of information gained about the \emph{true} causal orientation for a single intervention will therefore scale as $\log(N)$: the more we know about the system, the more information can be gained from a single intervention. 

A more important question is how this expression behaves when $n$ is randomly generated. We examine this by sampling $1000$ realizations of $n$ using $P$, and use these to compute the average realized gain and expected gain, \cref{eq:12} and \cref{eq:9} respectively.  We plot these along with the Bayesian update, which we obtain using the true probabilities as in \cref{eq:1} (see \cref{fig3} (a)). The simulations use $\alpha = 2$ and $\rho = 0.9$. The shaded area corresponds to one standard deviation; by symmetry both $\hat x$ and $\hat y$ behave the same. The realized gain is in this case larger than the expected gain due to the prior term $\alpha$. This can be understood by noting that the Jeffrey divergence, due to the $\log$-term, is quite unstable for low probability events. This means that the prior for low-probability events can generally be expected to play a large role in causal inference, and an under-estimation in particular may lead to very over-confident updates. We return to this point in Example 3. 

\begin{figure}[t!]
	\centering
	\includegraphics[width=.8\linewidth]{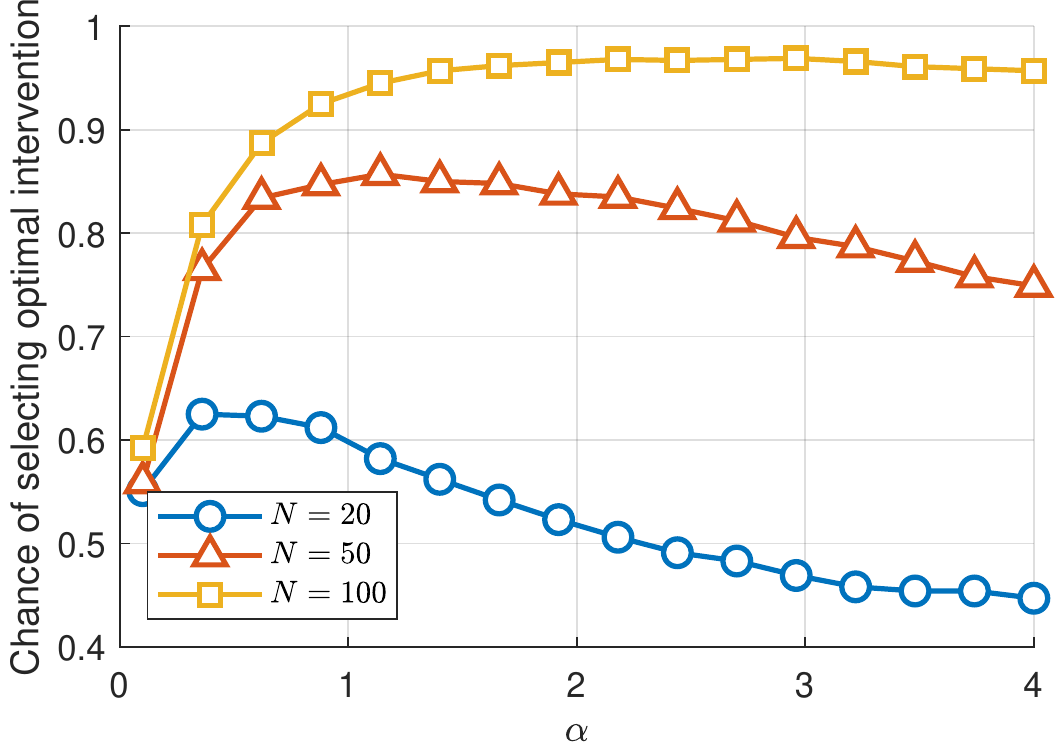}
	\caption{The problem described in Example 3, shown in \cref{fig3} (right), but where we consider the chance of selecting the single best intervention ($\hat x=1$) as a function of prior $\alpha$ and for different number of observations. The prior is necessary to obtain stable estimates of the expected information gain, however, it will impact the estimate of the expected information gain slightly differently and therefore a large prior may change the ordering. The method selects the best intervention with a much higher probability than chance even for very low counts. }\label{fig4}
\end{figure}

\paragraph{Example 2: Exploration}
This example illustrates how the method can guide exploration. Assume $K_X = K_Y = 4$ and $P_{11} = \rho$ and otherwise $P_{xy} = \frac{1-\rho}{ K^2-1}$. By symmetry, we only need to consider $\doo(X=1)$ and $\doo(X=2)$. The result, using the same settings as in Example 1, can be found in \cref{fig3} (b). Although $\hat x =1$ is by far the most likely event, it is not informative about the causal orientation, since $P(y|x=1)$ is very nearly equal to $P(y)$. Note that in this case the expected gain is larger than the realized gain. Interestingly, with as little as 20 samples, the method will suggest an optimal intervention.

\paragraph{Example 3: A single good intervention}
To illustrate a case where a single intervention is better than all the other, consider a problem where $P_{1,4} = P_{x,1} = \frac{\rho}{5}$ and otherwise $P_{xy} = \frac{1-\rho}{11}$. Both realized and expected gain of all interventions (that are not similar by symmetry) are visualized in \cref{fig3} (c), and $\hat x = 1$ constitutes the (single) optimal intervention (the standard deviations are not shown for visual clarity). %The variables are nearly independent so more samples have been included to show the expected gain eventually converge to the mean-field value. 
 Since the variables are nearly independent, more samples have been included, in order to show that the expected  gain eventually converges to the mean-field value.

We note that the expected information gain can be over-estimated in the small-sample limit (see e.g. $\hat y \neq 1$) where the prior term $\alpha$ will be more important.
 To gain more insight into this, we consider the same example, but now show the probability that the expected gain will be the largest for the optimal intervention $\hat x=1$, i.e. the chance that the agent will actually select the optimal intervention (see  \cref{fig4}). This probability is plotted as a function of the regularization parameter $\alpha$ for three representative numbers of samples. Although the ability to select the optimal intervention is more impacted by the prior in the small-sample limit, we note that even for just 20 samples, it is much higher than chance ($\frac{1}{8}$). 

\paragraph{Example 4: Active learning}
This example will consider a concrete Active learning setting and show that Bayesian causal induction can learn quicker when actions are selected based on \cref{eq:9}, compared to random selection. We consider the ground-truth as fixed at $H_0 = h^0_{X \rightarrow Y}$, and generate larger random problems by setting $K_X = K_Y  =8$ and setting $p_{xy} = \frac{u_{xy}}{\sum_{x,y} u_{xy}}$, where $u_{xy}$ are i.i.d. uniform random variables in $[0,1]$. Given $P_{xy}$, we sample $n$ as in the previous examples, and compute the information gain in favor of of $h^0_{X\rightarrow Y}$ based on an intervention $\hat x$ or $\hat y$ and corresponding observations of $y$ and $x$ using the Bayesian update given in \cref{eq:6}. We then consider the case where interventions are selected randomly, 
as well as the case where
 they are selected using the maximal expected information gain computed using \cref{eq:9}. The results are averaged over $10^5$ simulations using $\alpha=2$. In both cases, Bayesian causal induction gains information about the true causal orientation, however, the informative action selections result in about twice as large gain in evidence on average.

\begin{figure}[t!]
	\centering
	\includegraphics[width=.8\linewidth]{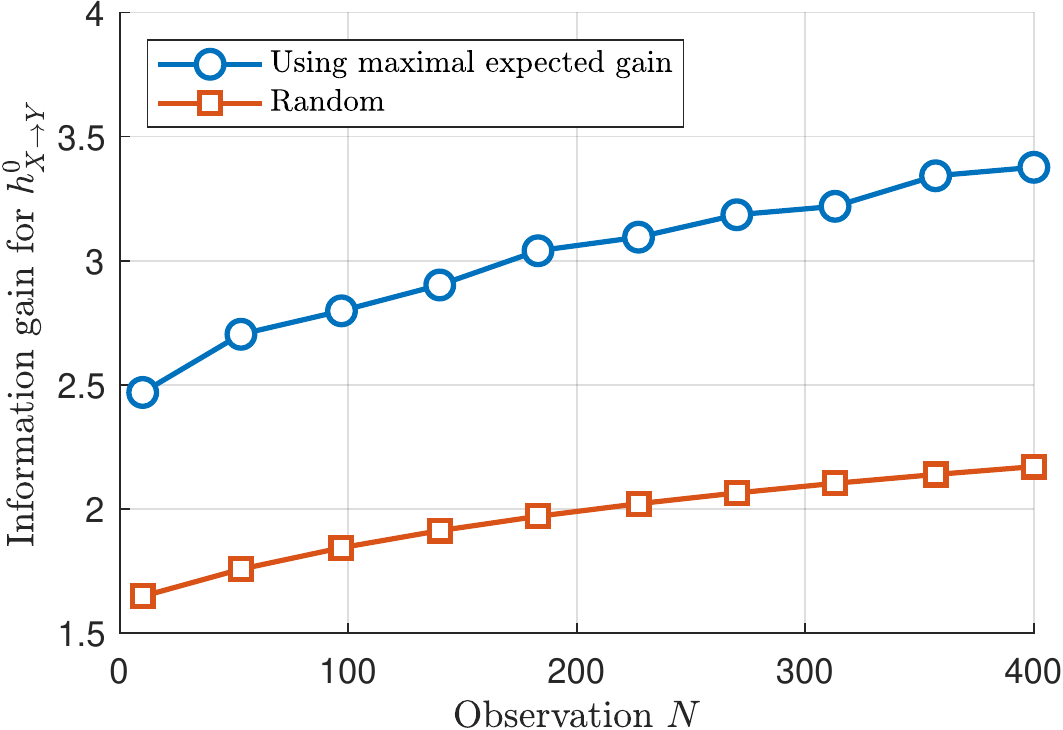}
	\caption{Evaluations of method for an actual intervention-selection problem. The truth is considered fixed as $H^0 = h^0_{X \rightarrow Y}$. From this, $10^5$ random joint distributions $P(X,Y)$ are generated (see text), and the information gains towards $h^0_{X \rightarrow Y}$ are computed when interventions are either selected randomly, or when using the maximum anticipated gain \cref{eq:9}. As shown, the information gain is in both cases positive, but about twice as large when interventions are selected using our method. }\label{fig5}
\end{figure}

\section{Discussion and conclusion}
Probability trees are conceptually the simplest possible approach to causal inference: consider the causal orientation as an event, specify a prior, and compute the posterior. Although recent work has demonstrated how probability trees can represent both interventions and counterfactual, as well as context-dependent causal statements that cannot be expressed in a directed acyclic causal model~\cite{genewein2020algorithms}, their practical use has remained limited.

In this work, we have highlighted another aspect of probability trees and Bayesian causal induction, namely the ability to make predictions about the information gain \emph{prior} to performing interventions. We have illustrated this in the simplest possible situation, and shown how to express both the the expected gain \emph{before} making an intervention, and the realized gain \emph{after} making an intervention. %, without committing to a ground-truth causal orientation $H^0$. 

In making statements such as this, it is important to emphasize that a reduction in uncertainty is by itself meaningless, rather, what matters is the information gain in favor of the true hypothesis, a distinction we make using $H^0$ and $H^a$.

In experiments, we have shown these measures can quantify the information gain and distinguish between different interventions. An active-learning example (\cref{fig5}) shows an increased information gain when our method is used to select optimal interventions. %. Our experiments show the relationship must depend on $\alpha$ and one quantity does not dominate the other. 

Many interesting avenues remain unexplored, such as the generalization to larger graphs, and concrete concentration bounds on the expected and realized information gains in terms of $P$.

\bibliography{library2}

\end{document}